\documentclass[11pt]{article}

\usepackage[preprint]{acl}

\usepackage{times}
\usepackage{latexsym}

\usepackage[T1]{fontenc}

\usepackage[utf8]{inputenc}

\usepackage{microtype}

\usepackage{inconsolata}

\usepackage{graphicx}

\usepackage{amsmath}  
\usepackage{amssymb}
\usepackage{mathtools}
\usepackage{array}
\usepackage{multirow} 
\usepackage{amsfonts}
\usepackage{booktabs} 
\usepackage{enumitem} 

\title{CRePE: Convolution-aware Relative Importance in\\Post-training Pruning with Efficient Search}

\author{Cheonjun Park \\
  Hankuk University of Foreign Studies \\
  \texttt{jun@hufs.ac.kr} \\}

\begin{document}
\maketitle

\begin{abstract}

Deploying Large Language Models (LLMs) in practice incurs substantial memory and computational costs. Post-training pruning (PTP) is an effective approach to reducing these costs by removing weights without additional training.
Among existing methods, RIA introduces relative importance scores normalized by row and column sums, achieving state-of-the-art accuracy.
However, RIA considers only 1D cross-shaped (row/column) directional information and assigns equal weight to row and column contributions.
In this paper, we propose \textbf{CRePE}, which incorporates 2D local neighborhood context and adaptive coefficients into Relative Importance scoring. CRePE consistently outperforms existing PTP methods across diverse models and sparsity settings. 
However, identifying optimal adaptive coefficients via perplexity (PPL)-based hill climbing requires numerous PPL evaluations and approximately 11 hours of search time. To address this, we propose \textbf{PHO} (Proxy-based Hyperparameter Optimization), which eliminates the need for repeated PPL measurements and reduces the search time to approximately 20 minutes. 
Furthermore, the optimal hyperparameter configuration found by PHO on one model transfers well to other models, demonstrating strong generalization.
Finally, we verify that CRePE can be orthogonally combined with existing techniques including Channel Permutation, non-uniform sparsity allocation, and re-pruning methods.

\end{abstract}


\begin{figure*}[t]
    \centering
    \includegraphics[width=\linewidth]{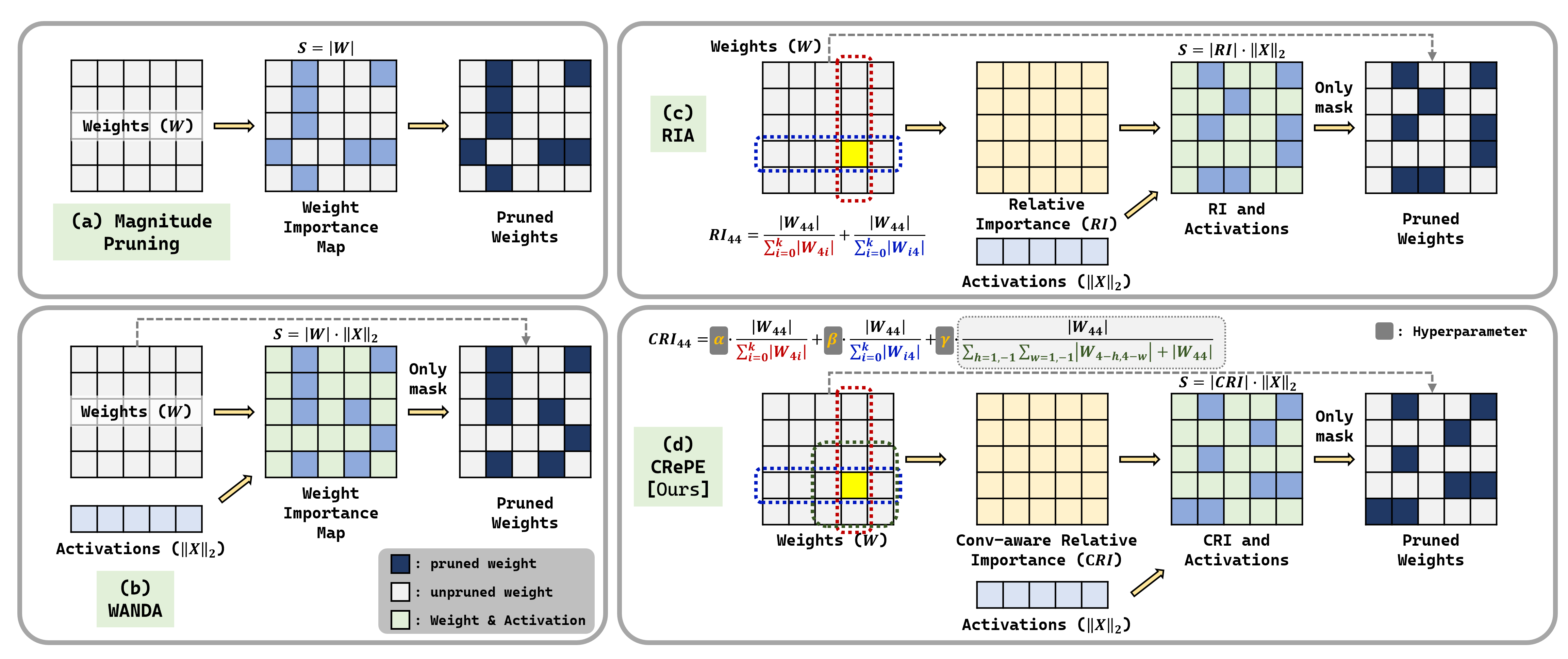}
    \caption{Importance score computation of (a) Magnitude Pruning, (b) Wanda, (c) RIA, and (d) CRePE. CRePE extends RIA's cross-pattern with a 2D neighborhood term (Sec~\ref{sec:2d}) and adaptive coefficients $\alpha$, $\beta$, $\gamma$ (Sec~\ref{sec:adaptive}).}
    \label{fig:method}
\end{figure*}

\section{Introduction}
\label{sec:intro}

Large Language Models (LLMs) have demonstrated remarkable performance across a wide range of natural language processing tasks~\cite{brown2020language, touvron2023llama}, yet their billions of parameters impose substantial memory and computational costs during inference. To address this, various compression techniques have been explored, including quantization~\cite{frantar2022gptq, xiao2023smoothquant}, knowledge distillation~\cite{hinton2015distilling}, and weight pruning~\cite{han2015learning,sun2023simple}. Among these, post-training pruning (PTP) removes unnecessary weights without any retraining, making it a practical and training-free solution well-suited for LLMs.

Among PTP methods, SparseGPT~\cite{frantar2023sparsegpt} first demonstrated the viability of LLM pruning by leveraging Hessian approximations to jointly select and update weights. 
Wanda~\cite{sun2023simple} introduced an importance criterion based on the product of weight magnitudes and input activations, achieving competitive performance without weight updates. 
RIA~\cite{zhang2024plug} further advanced this line of work by introducing Relative Importance (RI), which normalizes each weight by the sum of its row and the sum of its column.
Unlike weight update-based methods such as SparseGPT, which require approximately 90 minutes to prune LLaMA-2-70B, RIA achieves competitive accuracy through scoring alone, reducing pruning time to approximately 10 minutes.

Despite its efficiency, RIA has two fundamental limitations. 
\textbf{First}, RIA's importance scores rely solely on 1D cross-shaped directional information from the weight matrix. As weight matrices are inherently 2D structures, diagonal neighboring weights can also contribute to RI estimation. 
\textbf{Second}, RIA treats row and column contributions equally. Our preliminary experiments show that row and column directions contribute asymmetrically to pruning quality: using row-wise RI alone yields lower perplexity than column-wise RI alone (PPL 6.83 vs.\ 7.49 on LLaMA-2-7B), making uniform treatment suboptimal.

Motivated by these observations, we propose \textbf{CRePE} (\textbf{C}onvolution-aware \textbf{Re}lative Importance in \textbf{P}ost-training Pruning with \textbf{E}fficient Search), which extends RIA in two ways. First, it augments the cross-shaped row and column terms with a 2D neighborhood term to capture local weight context. Second, it assigns adaptive coefficients  $\alpha$, $\beta$, $\gamma$ to the row, column, and neighborhood contributions, respectively, to explicitly model their asymmetric importance. CRePE consistently outperforms existing PTP methods across diverse models and sparsity settings.

However, CRePE introduces a new challenge: finding the optimal combination of hyperparameters (kernel size $k$ and adaptive coefficients $\alpha$, $\beta$, $\gamma$). A naive PPL-based hill climbing search requires approximately 10 hours even on LLaMA-2-7B, undermining the efficiency advantage of PTP. 
To address this, we propose \textbf{PHO} (\textbf{P}roxy-based \textbf{H}yperparameter \textbf{O}ptimization). 
We first conduct a systematic Spearman correlation analysis across diverse hyperparameter configurations and discover that the Gini coefficient~\cite{ceriani2012origins} of the pruned weight importance scores at the first-layer \texttt{q\_proj} exhibits a strong positive correlation with PPL ($\rho = 0.953$). 
This finding suggests that Gini\textsubscript{pruned} can serve as a reliable surrogate for PPL evaluation. PHO leverages this proxy within a CMA-ES-based search framework, which enables efficient exploration of the hyperparameter space without repeated PPL measurements, reducing total search time from 10 hours to approximately 20 minutes — a $\mathbf{30\times}$ \textbf{speedup}.

The main contributions are as follows:
\begin{itemize}[noitemsep, topsep=0pt]
    
    \item We propose CRePE, which extends RIA's 1D cross-shaped Relative Importance with a 2D neighborhood term (Section 3.2) and adaptive coefficients (Section 3.3) to explicitly model the differential contributions of row, column, and neighborhood terms.
    
    \item We discover that the Gini coefficient of pruned weight importance scores strongly correlates with PPL ($\rho = 0.953$), and leverage this as a proxy objective in PHO, an efficient CMA-ES-based hyperparameter optimization framework that reduces search time by $30\times$ speedup over PPL-based hill climbing.
    
    \item We demonstrate consistent improvements over baseline PTP methods across diverse models including LLaMA-1/2/3, Qwen-2.5/3/3.5, Phi-4, and DeepSeek under unstructured, 2:4, and 4:8 sparsity settings.
    
    \item We empirically verify that CRePE can be orthogonally combined with existing pruning extensions, including Channel Permutation, non-uniform sparsity allocation (OWL, AlphaPruning, DLP), and re-pruning (DSnoT).
    
\end{itemize}

\section{Related Work}
\label{sec:formatting}

\subsection{Post-training Pruning}

Post-training pruning (PTP) removes weights from pretrained models without retraining, making it particularly attractive for LLMs where fine-tuning at scale is prohibitively expensive. SparseGPT~\cite{frantar2023sparsegpt} was the first practical PTP method for LLMs, framing pruning as a layerwise reconstruction problem and leveraging approximate Hessian information to jointly select and update weights. Wanda~\cite{sun2023simple} introduced an importance criterion based on the product of weight magnitudes and the $\ell_2$ norm of input activations, achieving performance close to SparseGPT without any weight updates. RIA~\cite{zhang2024plug} further improved upon this by normalizing each weight by its row and column sums to obtain a Relative Importance score, mitigating the channel corruption problem. CRePE belongs to the same scoring-only category as Wanda and RIA, requiring neither retraining nor weight updates~\citep{zhao2025fistapruner,meng2024alps,bovza2024fast,frantar2023sparsegpt}.
In addition to unstructured and weight update-based methods, structured pruning~\citep{ma2023llm, ashkboos2024slicegpt, men2024shortgpt, an2024fluctuation} and layer-level pruning~\cite{song2024sleb,yang2024laco,men2024shortgpt} remove entire model components such as attention heads, layers, or neurons, yielding hardware-friendly models without requiring specialized sparse kernels.

\subsection{Non-uniform LLM Pruning}

Most existing LLM pruning methods apply a uniform sparsity ratio across all layers. OWL~\cite{yin2024owl} discovered that the layerwise outlier distribution in LLMs is highly non-uniform, and proposed a non-uniform sparsity allocation strategy that assigns lower sparsity to layers with more outliers. AlphaPruning~\cite{lu2024alphapruning} estimates layerwise importance using Random Matrix Theory to guide non-uniform allocation. DLP~\cite{chen2025dlp} proposes dynamic layer prioritization via task-specific calibration. These methods focus on \emph{how much} to prune each layer, whereas CRePE focuses on \emph{which weights} to prune within each layer via improved importance scoring. The two approaches are orthogonal, and we experimentally verify that CRePE can be effectively combined with OWL, AlphaPruning, and DLP.
\section{CRePE}
\label{sec:crepe}

\paragraph{Preliminaries}
Given a pretrained weight matrix $\mathbf{W} \in \mathbb{R}^{C_\text{out} \times C_\text{in}}$ and the corresponding input activation $\mathbf{X} \in \mathbb{R}^{C_\text{in} \times T}$, the goal of post-training pruning is to find a binary mask $\mathbf{M} \in \{0, 1\}^{C_\text{out} \times C_\text{in}}$ that zeros out a target fraction of weights while minimizing the degradation in model performance. Here, $X_j \in \mathbb{R}^T$ denotes the $j$-th column of $\mathbf{X}$, representing the input activation of the $j$-th input channel.

\begin{figure}[hbt!]
    \centering
    \includegraphics[width=\linewidth]{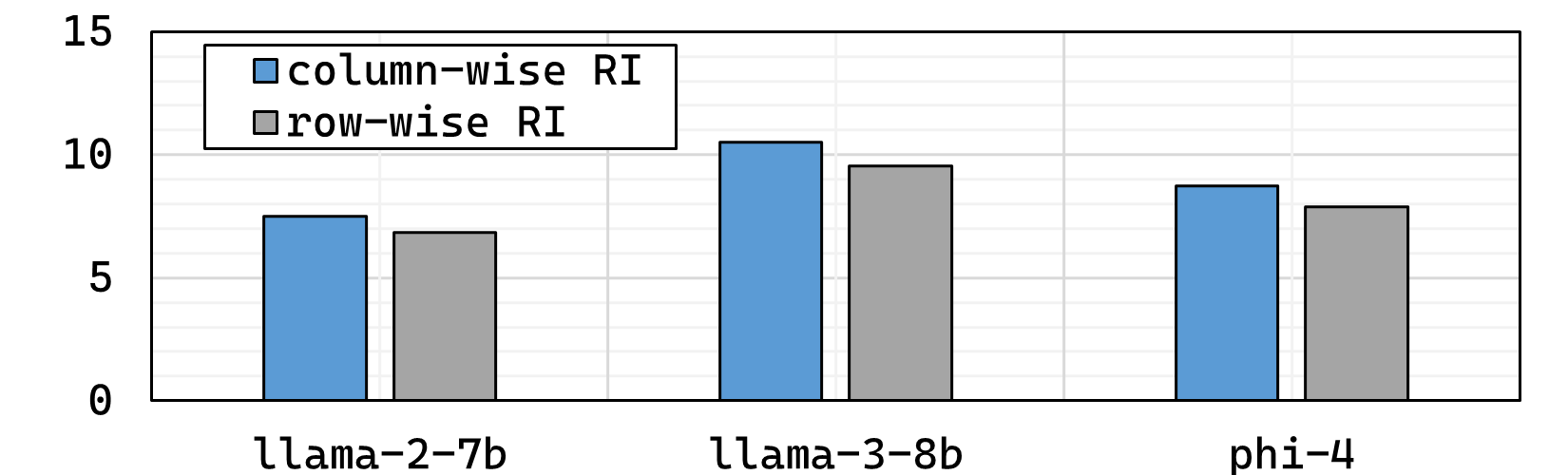}
    \caption{Ablation on directional contributions. PPL under unstructured sparsity 50\%. Column-only and row-only refer to using $\beta$ or $\alpha$ exclusively, respectively.}
    \label{fig:direction}
\end{figure}

\subsection{Motivation}
The importance score of RIA is defined as:
\begin{equation}
S_{ij}^{\text{RIA}} = \left( \frac{|W_{ij}|}{\sum_k |W_{ik}|} + \frac{|W_{ij}|}{\sum_k |W_{kj}|} \right) \cdot \|X_j\|_2
\label{eq:ria_score}
\end{equation}
As illustrated in Figure~\ref{fig:method}(c) and Equation ~\ref{eq:ria_score}, RIA evaluates the importance of each weight $W_{ij}$ using only the 1D cross-shaped directional information from its row and column sums. However, weight matrices are inherently 2D structures, and diagonal neighboring weights can also contribute to importance estimation. Furthermore, RIA treats the row and column directions equally, despite the fact that their impacts on pruning quality differ in practice. 
As shown in Figure~\ref{fig:direction}, using row-wise RI alone yields lower perplexity than column-wise RI alone on LLaMA-2-7B (6.83 vs.\ 7.49), indicating that the two directions contribute asymmetrically to pruning quality. Treating them equally, as RIA does, is therefore suboptimal.
In the following, we propose \textbf{Convolution-aware Relative Importance (CRI)}, which addresses these two limitations by integrating a 2D neighborhood term (Section~\ref{sec:2d}) and adaptive per-term coefficients (Section~\ref{sec:adaptive}).

\subsection{2D Neighborhood Relative Importance}
\label{sec:2d}
To extend the Relative Importance score with 2D structural information, we define a neighborhood term $N_{ij}^{(k)}$ that captures the local 2D context around each weight $W_{ij}$ while excluding the cross-shaped neighbors already captured by the RIA terms:
\begin{equation}
N_{ij}^{(k)} = \sum_{(h,w) \in \mathcal{D}_k} |W_{i+h,\, j+w}|
\label{eq:neighborhood}
\end{equation}
where $\mathcal{D}_k = \{(h, w) : |h|, |w| \leq \lfloor k/2 \rfloor,\, hw \neq 0\} \cup \{(0,0)\}$. The condition $hw \neq 0$ excludes the cross pattern (row and column neighbors), while $(0,0)$ retains the self-magnitude. Note that $(0,0) \in \mathcal{D}_k$ ensures that $|W_{ij}|$ itself contributes to the denominator of the neighborhood term, making it analogous in form to the row and column normalization in Equation~\ref{eq:ria_score}. For example, $\mathcal{D}_3 = \{(0,0),\,(\pm1,\pm1)\}$ contains $5$ elements, and $\mathcal{D}_5$ contains $17$.
This neighborhood sum can be efficiently computed via 2D convolution of $|\mathbf{W}|$ with a masked kernel, justifying the ``convolution-aware'' in our method name.

\subsection{Adaptive Relative Importance}
\label{sec:adaptive}
The row, column, and neighborhood terms may contribute differently to the importance of a weight. To explicitly model these differential contributions, we introduce per-term coefficients $\alpha$, $\beta$, and $\gamma$ that weight the three terms:
\begin{equation}
\text{CRI}_{ij} = \alpha \cdot \frac{|W_{ij}|}{\sum_k |W_{ik}|} + \beta \cdot \frac{|W_{ij}|}{\sum_k |W_{kj}|} + \gamma \cdot \frac{|W_{ij}|}{N_{ij}^{(k)}}
\label{eq:cri}
\end{equation}
where $\alpha$, $\beta$, and $\gamma$ control the contributions of the row, column, and 2D neighborhood terms, respectively, and are determined via PHO (Section~\ref{sec:pho}). The final CRePE importance score is obtained by combining CRI with the $\ell_2$ norm of the input activation:
\begin{equation}
S_{ij}^{\text{CRePE}} = \text{CRI}_{ij} \cdot \|X_j\|_2
\label{eq:CRePE_score}
\end{equation}
When $\alpha = \beta = 1$ and $\gamma = 0$, CRePE reduces to RIA, confirming that RIA is a special case of CRePE.

\begin{figure*}[t]
    \centering
    \includegraphics[width=\linewidth]{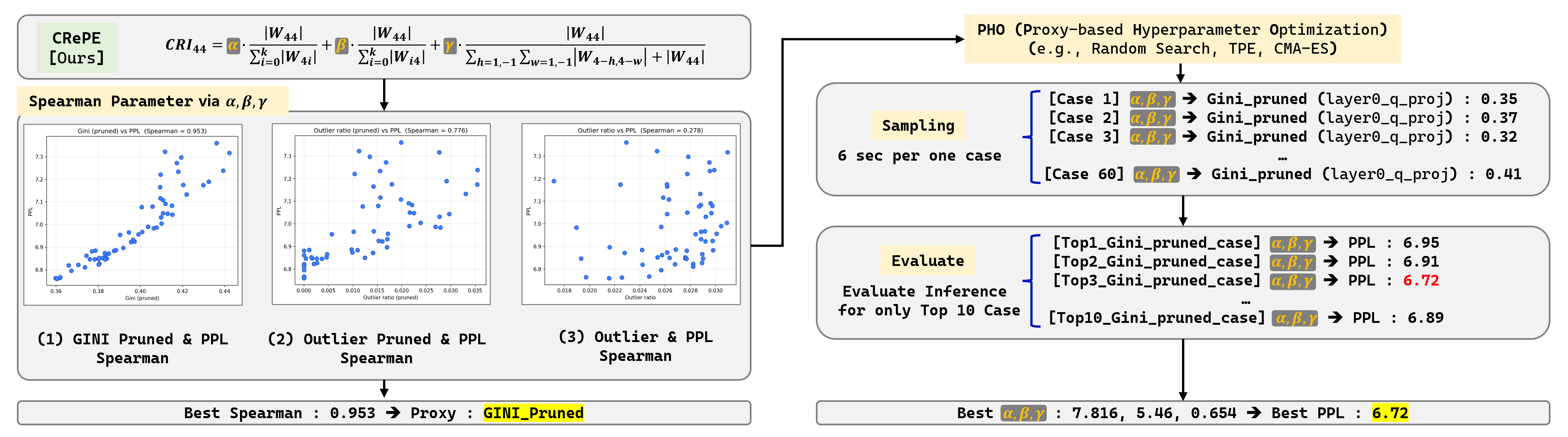}
    \caption{Overview of PHO (Proxy-based Hyperparameter Optimization). \textbf{Left}: Spearman correlation analysis across three proxy candidates identifies Gini\textsubscript{pruned} as the most reliable surrogate ($\rho=0.953$). \textbf{Right}: Two-stage search procedure — CMA-ES samples 60 candidates evaluated via proxy ($\sim$6 sec each), followed by full PPL evaluation on the top 10 candidates ($\sim$90 sec each).}
    \label{fig:pho}
\end{figure*}

\section{PHO}
\label{sec:pho}

\begin{table}[t]
\centering
\caption{Comparison of search algorithms in PHO on LLaMA-2-7B across unstructured, 2:4, and 4:8 sparsity. PPL and search time are reported. Hill climbing serves as the upper-bound reference.}
\label{tab:search}
\resizebox{\linewidth}{!}{%
\begin{tabular}{lcccccc}
\toprule
\multirow{3}{*}{\textbf{Method}} & \multicolumn{6}{c}{\textbf{LLaMA-2-7B}} \\
\cmidrule(lr){2-7}
 & \multicolumn{2}{c}{\textbf{Unstructured (50\%)}} & \multicolumn{2}{c}{\textbf{2:4}} & \multicolumn{2}{c}{\textbf{4:8}} \\
\cmidrule(lr){2-3} \cmidrule(lr){4-5} \cmidrule(lr){6-7}
 & \textbf{PPL} & \textbf{Time} & \textbf{PPL} & \textbf{Time} & \textbf{PPL} & \textbf{Time} \\
\midrule
RIA                & 6.81   & 107s      & 11.29   & 111s     & 8.38   & 109s     \\
CRePE (Hill Climbing)       & 6.74   & 10.9h      & 10.74   & 10.8h     & 8.16   & 11.0h     \\
PHO (TPE)     & 6.77   & 1213s  & 10.81   & 1311s & 8.17   & 1255s \\
PHO (Random)  & 6.76   & 1213s  & 10.86   & 1309s & 8.18   & 1258s  \\
PHO (CMA-ES)  & 6.75   & 1212s  & 10.83   & 1313s & 8.16   & 1255s \\
\bottomrule
\end{tabular}%
}
\end{table}

\subsection{Motivation}
CRePE introduced in Section~\ref{sec:crepe} has hyperparameters consisting of the kernel size $k$ and adaptive coefficients $(\alpha, \beta, \gamma)$. Finding their optimal combination via PPL-based hill climbing requires full model pruning and PPL evaluation for each candidate, taking approximately 11 hours on LLaMA-2-7B (see Table~\ref{tab:search}), compared to only $\sim$110 seconds for RIA and Wanda. This substantially undermines the key advantage of post-training pruning — fast, training-free applicability. To address this, we propose \textbf{PHO} (Proxy-based Hyperparameter Optimization), a proxy-guided search framework that predicts hyperparameter quality without directly measuring PPL.

\begin{table}[t]
\centering
\caption{Spearman correlation between candidate proxy metrics and PPL across three models. Higher absolute value indicates stronger predictive power.}
\label{tab:proxy}
\resizebox{\linewidth}{!}{%
\begin{tabular}{lccc}
\toprule
\textbf{Metric} & \textbf{Qwen3.5-0.8B} & \textbf{LLaMA-2-7B} & \textbf{LLaMA-2-13B} \\
\midrule
\textbf{Outlier vs. PPL}                  &  0.52   &  0.28   &  0.21  \\
\textbf{Outlier (retained) vs. PPL}       & -0.27   & -0.05   &  0.22  \\
\textbf{Outlier (pruned) vs. PPL}         &  0.17   &  0.78   &  0.71  \\
\textbf{Gini (retained) vs. PPL}        &  0.90   &  0.72   &  0.66  \\
\textbf{Gini (pruned) vs. PPL}           &  \textbf{0.95}   &  \textbf{0.95}   &  \textbf{0.92}  \\
\bottomrule
\end{tabular}
}
\end{table}

\subsection{Proxy Definition}
An effective proxy must satisfy two conditions: (1) strong correlation with PPL, and (2) low computational cost. Rather than pruning the full model, we prune only the \texttt{q\_proj} module of the first transformer layer and extract the proxy from the distribution of pruned weight importance scores. 
Since \texttt{q\_proj} at the first layer is the first module processed in the pruning pipeline, this single-module evaluation reduces the number of pruned modules by approximately 200$\times$ compared to full model pruning.

The Gini coefficient takes values in $[0, 1]$, where 0 indicates a perfectly uniform distribution and 1 indicates maximal inequality. A high Gini\textsubscript{pruned} indicates that the importance score distribution within the pruned set forms a long tail, meaning that a small number of pruned weights carry disproportionately high scores relative to the others. Under well-balanced $(\alpha, \beta, \gamma)$, pruned weights share uniformly low scores (low Gini), indicating that the pruning boundary cleanly separates low- from high-scoring weights. Conversely, ill-balanced $(\alpha, \beta, \gamma)$ distort the normalization, causing weights with disproportionately high scores to enter the pruned set, which increases the layer-wise reconstruction error and ultimately degrades PPL.

We measure various proxy candidates at the first-layer \texttt{q\_proj} and analyze their Spearman correlation with PPL across three models (Qwen3.5-0.8B, LLaMA-2-7B/13B). As shown in Table~\ref{tab:proxy}, Gini\textsubscript{pruned} exhibits the strongest and most consistent positive correlation with PPL among all candidates, outperforming Outlier\textsubscript{pruned} and Gini\textsubscript{retained}.

\subsection{Search Algorithm}
\paragraph{Analysis}
Various optimization algorithms can be applied to minimize Gini\textsubscript{pruned} as the surrogate objective. We compare three representative algorithms: CMA-ES (Covariance 
Matrix Adaptation Evolution Strategy), TPE (Tree-structured Parzen Estimator), and Random Search. As shown in Table~\ref{tab:search}, all three algorithms achieve comparable PPL within 0.03 of each other, while CMA-ES performs best on average and is particularly well-suited to CRePE's continuous hyperparameter space $(\alpha, \beta, \gamma \in \mathbb{R})$ via its covariance adaptation mechanism. We therefore adopt CMA-ES as the default sampler.

\paragraph{Overall Procedure of PHO}
As illustrated in Figure~\ref{fig:pho}, the full PHO search procedure consists of two stages (reported for LLaMA-2-7B):
\begin{itemize}[noitemsep, topsep=0pt]
    \item \textbf{Stage 1 --- Proxy-based screening}: A CMA-ES sampler generates 60 hyperparameter candidates. For each candidate, Gini\textsubscript{pruned} is computed from the \texttt{q\_proj} of layer 0. Each evaluation takes approximately 6 seconds.
    \item \textbf{Stage 2 --- PPL-based verification}: Full model pruning and WikiText2 PPL evaluation are performed only for the top 10 candidates with the lowest Gini\textsubscript{pruned}. Each evaluation takes approximately 90 seconds.
\end{itemize}
The total search time is approximately 20 minutes, representing a speedup of more than $30\times$ over PPL-based hill climbing ($\sim$11 hours). The number of trials and top-$k$ candidates are set to 60 and 10, respectively, based on empirical validation; further analysis is provided in Section~\ref{sec:ablation_search}.

\section{Experiments}

\subsection{Experimental Setup}
\textbf{Models and datasets.}
We evaluate on a diverse set of open-source LLMs including LLaMA (7B--65B)~\cite{touvron2023llama}, LLaMA-2 (7B--70B)~\cite{touvron2023llama2openfoundation}, LLaMA-3 (8B)~\cite{grattafiori2024llama3herdmodels}, Qwen-2.5/3.5~\cite{qwen2025qwen25technicalreport,team2026qwen3}, Phi-4~\cite{abdin2024phi4technicalreport}, and DeepSeek-LLM~\cite{deepseekai2024deepseekllmscalingopensource}. We use the public checkpoints of the involved models in the HuggingFace Transformers library. Language modeling performance is measured by perplexity (PPL) on the WikiText2 dataset. Zero-shot performance is evaluated by average accuracy across six tasks: BoolQ, MNLI, RTE, HellaSwag, ARC-Easy, and ARC-Challenge. For calibration, we use 128 sequences of length 2048 randomly sampled from the C4 dataset.

\textbf{Implementation.}
Hyperparameter search for CRePE is performed via PHO with a CMA-ES sampler, generating 60 candidates in Stage 1 and evaluating the top 10 via full PPL measurement in Stage 2. 
Based on the ablation in Section~\ref{sec:ablation_kernel}, we use a kernel size of $k=9$ for all CRePE experiments unless otherwise specified.
All experiments are conducted on 2 NVIDIA RTX PRO 6000 Blackwell GPUs, each equipped with 96GB of memory.


\subsection{Main Results}

\begin{table*}[t]
\centering
\small
\caption{WikiText-2 perplexity results for unstructured (50\%) and 2:4 structured pruning across LLaMA model families. \textbf{Bold} values indicate the best (lowest) perplexity per column.}
\label{tab:pruning_results}
\setlength{\tabcolsep}{4pt}
\resizebox{\linewidth}{!}{%
\begin{tabular}{l|cccc|ccc|c|cccc|ccc|c}
\toprule
\multirow{3}{*}{\textbf{Method}} 
& \multicolumn{8}{c|}{\textbf{Unstructured (50\%)}} 
& \multicolumn{8}{c}{\textbf{2:4 Pruning}} \\
\cmidrule(lr){2-9} \cmidrule(lr){10-17}
& \multicolumn{4}{c|}{\textbf{LLaMA}} 
& \multicolumn{3}{c|}{\textbf{LLaMA-2}} 
& \textbf{LLaMA-3}
& \multicolumn{4}{c|}{\textbf{LLaMA}} 
& \multicolumn{3}{c|}{\textbf{LLaMA-2}} 
& \textbf{LLaMA-3} \\
\cmidrule(lr){2-5} \cmidrule(lr){6-8} \cmidrule(lr){9-9}
\cmidrule(lr){10-13} \cmidrule(lr){14-16} \cmidrule(lr){17-17}
& \textbf{7B} & \textbf{13B} & \textbf{30B} & \textbf{65B} 
& \textbf{7B} & \textbf{13B} & \textbf{70B} 
& \textbf{8B}
& \textbf{7B} & \textbf{13B} & \textbf{30B} & \textbf{65B} 
& \textbf{7B} & \textbf{13B} & \textbf{70B} 
& \textbf{8B} \\
\midrule
Dense     & 5.68 & 5.09 & 4.77 & 3.56 & 5.47 & 4.88 & 3.32 & 6.14
          & 5.68 & 5.09 & 4.77 & 3.56 & 5.47 & 4.88 & 3.32 & 6.14 \\
Magnitude & 17.28 & 20.22 & 7.54 & 5.90 & 16.02 & 6.83 & 5.36 & 205.46
          & 42.55 & 18.32 & 9.11 & 7.11 & 37.76 & 8.74 & 6.33 & 2401.90 \\
SparseGPT & 7.24 & 6.20 & 5.32 & 4.57 & 6.99 & 6.10 & 4.25 & 9.61
          & 11.54 & 9.01 & 7.14 & 6.25 & 11.01 & 8.77 & \textbf{5.16} & 16.31 \\
Wanda     & 7.26 & 6.15 & 5.24 & 4.57 & 6.92 & 5.99 & 4.22 & 9.40
          & 11.01 & 9.59 & 6.90 & 6.28 & 12.13 & 9.00 & 5.40 & 24.08 \\
RIA       & 7.12 & 6.08 & 5.08 & 4.38 & 6.81 & 5.83 & 4.11 & 9.34
          & 11.10 & 8.95 & 6.73 & 6.03 & 11.29 & 8.41 & 5.38 & 22.94 \\
\textbf{CRePE}     & \textbf{7.07} & \textbf{6.03} & \textbf{5.07} & \textbf{4.37} & \textbf{6.74} & \textbf{5.80} & \textbf{4.09} & \textbf{9.31}
          & \textbf{10.69} & \textbf{8.65} & \textbf{6.59} & \textbf{5.90} & \textbf{10.74} & \textbf{7.81} & 5.32 & \textbf{22.85} \\
\bottomrule
\end{tabular}
}
\end{table*}

\begin{table*}[hbt!]
\centering
\small
\caption{WikiText-2 perplexity across diverse model architectures under unstructured 50\% and 2:4 pruning settings. \textbf{Bold} indicates the best (lowest) perplexity among pruning methods.}
\label{tab:generalization}
\resizebox{\linewidth}{!}{%
\setlength{\tabcolsep}{5pt}
\begin{tabular}{l|cccc|cccc}
\toprule
\multirow{2}{*}{\textbf{Method}}
& \multicolumn{4}{c|}{\textbf{Unstructured 50\%}}
& \multicolumn{4}{c}{\textbf{2:4 Pruning}} \\
\cmidrule(lr){2-5} \cmidrule(lr){6-9}
& \textbf{Phi4-14B} & \textbf{Qwen3.5-9B} & \textbf{Qwen2.5-7B} & \textbf{DeepSeek-7B}
& \textbf{Phi4-14B} & \textbf{Qwen3.5-9B} & \textbf{Qwen2.5-7B} & \textbf{DeepSeek-7B} \\
\midrule
Dense     & 6.458  & 8.644    & 6.849   & 12.281  & 6.458  & 8.644   & 6.849    & 12.281  \\
\midrule
Magnitude & 11.939 & 1243.500 & 308.200 & 156.900 & 14.442 & 33.241  & 2485.100 & 301.800 \\
Wanda     & 8.485  & 10.671   & 9.040   & 20.853  & 10.837 & 15.995  & 14.799   & \textbf{30.136} \\
RIA       & 7.939  & 9.412    & 8.574   & 16.978  & 10.510 & 14.599  & 13.804   & 31.485  \\
\textbf{CRePE} & \textbf{7.891} & \textbf{9.403} & \textbf{8.526} & \textbf{16.806} & \textbf{10.453} & \textbf{14.335} & \textbf{13.722} & 31.118 \\
\bottomrule
\end{tabular}
}
\end{table*}

\subsubsection{Perplexity}

\paragraph{Unstructured 50\% Pruning.}
Table~\ref{tab:pruning_results} shows WikiText-2 perplexity results under the unstructured 50\% sparsity setting.
CRePE achieves lower perplexity than RIA across all evaluated models, with consistent improvements on LLaMA-7B (7.07 vs.\ 7.12) and LLaMA-2-7B (6.74 vs.\ 6.81).

\paragraph{2:4 Structured Pruning.}
The 2:4 sparsity pattern enables hardware acceleration via NVIDIA Sparse Tensor Cores, making it a practically relevant setting.
As shown in Table~\ref{tab:pruning_results}, CRePE achieves meaningful gains over RIA on LLaMA-2-7B (10.74 vs.\ 11.29) and LLaMA-2-13B (7.81 vs.\ 8.41).
These absolute improvements are larger than those observed under unstructured pruning, suggesting that more stringent structural constraints amplify the benefit of refined importance scoring.

\paragraph{Generalization Across Architectures.}
Table~\ref{tab:generalization} evaluates CRePE on Phi4-14B, Qwen3.5-9B, Qwen2.5-7B, and DeepSeek-7B under unstructured 50\% and 2:4 pruning settings.
CRePE consistently outperforms RIA across nearly all architectures and sparsity patterns, with the sole exception of DeepSeek-7B under 2:4 pruning, where Wanda achieves the lowest perplexity.
These results confirm that CRePE generalizes robustly across diverse architectures without any architecture-specific modification.


\begin{table*}[t]
\centering
\caption{Zero-shot accuracy (\%) under unstructured 50\% and 2:4 sparsity across six tasks. Higher value is better.}
\label{tab:zeroshot}
\resizebox{\linewidth}{!}{%
\begin{tabular}{llccccccc}
\toprule
\textbf{Model} & \textbf{Method} & \textbf{RTE} & \textbf{MNLI} & \textbf{HellaSwag} & \textbf{BoolQ} & \textbf{ARC-Easy} & \textbf{ARC-Challenge} & \textbf{Average} \\
\midrule
LLaMA-2-7B & Dense & 65.34 & 43.14 & 60.07 & 80.61 & 79.46 & 48.46 & 64.78 \\
\midrule
\multirow{5}{*}{\shortstack{LLaMA-2-7B \\ (unstructured)}}
& Magnitude & 55.96 & 39.59 & 54.42 & 57.65 & 70.58 & 38.31 & 55.38 \\
& Wanda     & 54.51 & 43.54 & 54.90 & 77.89 & 74.49 & 41.21 & 59.40 \\
& SparseGPT & 61.01 & 43.27 & 56.15 & 81.35 & 76.22 & 42.75 & 61.34 \\
& RIA       & 58.84 & 44.49 & 56.83 & 80.98 & 76.05 & 41.47 & 61.14 \\
& \textbf{CRePE} & \textbf{66.43} & \textbf{44.81} & \textbf{57.34} & \textbf{80.31} & \textbf{75.84} & \textbf{42.15} & \textbf{62.18} \\
\midrule
\multirow{5}{*}{\shortstack{LLaMA-2-7B \\ (2:4)}}
& Magnitude & 53.79 & 32.90 & 50.10 & 65.72 & 62.33 & 31.74 & 50.66 \\
& Wanda     & 55.96 & 37.29 & 46.53 & 75.87 & 68.73 & 34.98 & 53.93 \\
& SparseGPT & 56.68 & 39.94 & 48.23 & 80.86 & 69.95 & 36.26 & 55.82 \\
& RIA       & 56.32 & 37.23 & 47.93 & 77.00 & 69.61 & 34.90 & 54.71 \\
& \textbf{CRePE} & \textbf{57.76} & \textbf{39.11} & \textbf{50.46} & \textbf{77.80} & \textbf{69.74} & \textbf{36.52} & \textbf{56.03} \\
\bottomrule
\end{tabular}%
}
\end{table*}

\subsubsection{Zero-shot Performance}
Table~\ref{tab:zeroshot} presents zero-shot performance across six tasks on LLaMA-2-7B under unstructured 50\% and 2:4 sparsity settings. Under unstructured 50\% pruning, CRePE achieves the highest average accuracy of 62.18\%, outperforming RIA (61.14\%) by 1.04\%. Notably, CRePE achieves the highest RTE accuracy (66.43\%) among all pruning methods, outperforming RIA (58.84\%) by 7.59 \%. Under the 2:4 setting, CRePE (56.03\%) likewise surpasses RIA (54.71\%) by 1.32 percentage points, maintaining consistent gains across all six tasks. These results demonstrate that the performance gains of CRePE are not confined to perplexity metrics but are consistently maintained on downstream tasks.


\subsection{Ablation Studies}

\subsubsection{Effect of Kernel Size}
\label{sec:ablation_kernel}

Table~\ref{tab:ablation_kernel} reports the effect of varying the 2D neighborhood kernel size, ranging from $3{\times}3$ to $11{\times}11$. On LLaMA-2-7B under unstructured 50\% sparsity, a $9{\times}9$ kernel achieves the lowest perplexity of 6.753, a clear improvement over RIA (6.810), while performance slightly degrades at $11{\times}11$ (6.754). The same trend is observed on Qwen3.5-0.8B, suggesting that a moderately sized local neighborhood is optimal and excessively large receptive fields introduce noise.

\begin{table}[t]
\centering
\caption{Effect of neighborhood kernel size $k$ on PPL.}
\label{tab:ablation_kernel}
\resizebox{\linewidth}{!}{%
\begin{tabular}{lcc}
\toprule
\multirow{2}{*}{\textbf{Method}} & \multicolumn{2}{c}{\textbf{Unstructured 50\%}} \\
\cmidrule(lr){2-3}
 & \textbf{Qwen3.5-0.8B} & \textbf{LLaMA-2-7B} \\
\midrule
Wanda                  & 59.041 & 6.922 \\
SparseGPT              & 35.903 & 6.991 \\
RIA                    & 35.212 & 6.810 \\
CRePE w/o kernel         & 34.997 & 6.804 \\
CRePE w/ 3$\times$3      & 34.956 & 6.804 \\
CRePE w/ 5$\times$5      & 34.529 & 6.757 \\
CRePE w/ 7$\times$7      & 34.492 & 6.756 \\
CRePE w/ 9$\times$9      & 34.492 & 6.753 \\
CRePE w/ 11$\times$11    & 34.656 & 6.754 \\
\bottomrule
\end{tabular}
}
\end{table}


\subsubsection{Analysis of PHO}
\label{sec:ablation_search}


Table~\ref{tab:cmaes_ablation} presents the effect of varying \texttt{n\_trials} and \texttt{top\_k} in CMA-ES-based PHO. Increasing \texttt{n\_trials} from 15 to 60 yields steady improvement (6.7791 $\to$ 6.7537), while the marginal gain diminishes beyond 60 trials (6.7537 $\to$ 6.7566), suggesting sufficient convergence within 60 trials. For \texttt{top\_k}, increasing from 5 to 10 provides a modest gain (6.7578 $\to$ 6.7537), while a further increase to 15 yields no additional benefit. Overall, (\texttt{n\_trials}$=60$, \texttt{top\_k}$=10$) offers the best trade-off, while (\texttt{n\_trials}$=30$, \texttt{top\_k}$=5$) achieves competitive perplexity (6.7645) within 10 minutes for time-constrained scenarios.

\begin{table}[t]
\centering
\small
\caption{Analysis of PHO. PPL on LLaMA-2-7B with CRePE 50\% unstructured pruning using CMA-ES.}
\label{tab:cmaes_ablation}
\begin{tabular}{cccc}
\toprule
\textbf{n\_trials} & \textbf{top\_k} & \textbf{Best PPL} & \textbf{time} \\
\midrule
15 &  5 & 6.7791 & 515.0s  \\
30 &  5 & 6.7645 & 607.9s  \\
30 & 10 & 6.7645 & 1026.8s \\
60 &  5 & 6.7578 & 799.8s  \\
60 & 10 & \textbf{6.7537} & \textbf{1212.7s}  \\
60 & 15 & 6.7578 & 1636.7s \\
90 & 10 & 6.7566 & 1406.9s \\
90 & 15 & 6.7559 & 1825.6s \\
\bottomrule
\end{tabular}
\end{table}

\begin{table}[t]
\centering
\small
\caption{Cross-sparsity transferability of PHO hyperparameters on LLaMA-2-13B. Hyperparameters optimized on LLaMA-2-7B under each sparsity setting are directly applied to LLaMA-2-13B without re-tuning.}
\label{tab:cross_sparsity}
\resizebox{\linewidth}{!}{%
\begin{tabular}{ll|c|ccc}
\toprule
\multirow{2}{*}{\textbf{Model}} & \multirow{2}{*}{\textbf{Sparsity}} & \multirow{2}{*}{\textbf{RIA}} & \multicolumn{3}{c}{\textbf{L-2-7B Best Hyperparameter}} \\
\cmidrule(lr){4-6}
 & & & \textbf{un 50\%} & \textbf{2:4} & \textbf{4:8} \\
\midrule
\multirow{3}{*}{\textbf{L-2-13B}}
 & un 50\% & 5.83 & 5.82 & 5.81 & 5.82 \\
 & 2:4     & 8.41 & 8.09 & 8.03 & 8.08 \\
 & 4:8     & 6.74 & 6.65 & 6.62 & 6.63 \\
\bottomrule
\end{tabular}
}
\end{table}

\begin{table*}[t]
\centering
\caption{Effect of combining CRePE with Channel Permutation (CP) and LSA CP under 2:4 and 4:8 sparsity.}
\label{tab:cp}
\resizebox{\linewidth}{!}{%
\begin{tabular}{llccccccc}
\toprule
\textbf{Model} & \textbf{Method} & \textbf{Unstructured 50\%} & \textbf{2:4} & \textbf{2:4 + CP} & \textbf{2:4 + LSA} & \textbf{4:8} & \textbf{4:8 + CP} & \textbf{4:8 + LSA} \\
\midrule
\multirow{5}{*}{\shortstack{LLaMA-2-7B\\(Dense: 5.47)}}
 & Magnitude & 16.02 & 37.76 & 253.39 & 95.45 & 15.91 & 29.01 & 21.55 \\
 & SparseGPT & 6.99  & 11.01 & 10.43  & 9.97  & 8.48  & 8.14  & 8.11  \\
 & Wanda     & 6.92  & 12.13 & 11.65  & 11.12 & 8.60  & 8.39  & 8.20  \\
 & RIA       & 6.81  & 11.29 & 10.36  & 10.03 & 8.38  & 8.18  & 8.01  \\
 & \textbf{CRePE}     & \textbf{6.74}  & \textbf{10.74} & \textbf{10.20}  & \textbf{9.80}  & \textbf{8.16}  & \textbf{8.00}  & \textbf{7.85}  \\
\midrule
\multirow{5}{*}{\shortstack{LLaMA-2-13B\\(Dense: 4.88)}}
 & Magnitude & 6.83 & 8.74 & 8.89 & 8.87 & 7.32 & 7.31 & 7.16 \\
 & SparseGPT & 6.10 & 8.77 & 8.61 & 8.48 & 7.01 & 6.82 & 6.80 \\
 & Wanda     & 5.99 & 9.00 & 8.74 & 8.45 & 7.00 & 6.93 & 6.83 \\
 & RIA       & 5.83 & 8.41 & 8.12 & 7.67 & 6.74 & 6.62 & 6.53 \\
 & \textbf{CRePE}     & \textbf{5.80} & \textbf{7.81} & \textbf{7.67} & \textbf{7.35} & \textbf{6.59} & \textbf{6.58} & \textbf{6.45} \\
\bottomrule
\end{tabular}%
}
\end{table*}

\subsubsection{Hyper-parameter Reuse}
\label{sec:ablation_Reuse}


\paragraph{Pruning Efficiency.}
The importance score computation in CRePE extends RIA's cross-shaped term with a 2D neighborhood term, incurring only a single additional convolution operation.
Consequently, the wall-clock pruning time of CRePE is on par with that of RIA and Wanda.
As reported by \citet{zhang2024plug}, both RIA and Wanda prune LLaMA-2-70B in approximately 10 minutes, whereas SparseGPT, which requires Hessian computation, takes around 90 minutes.
CRePE inherits the same efficiency as these scoring-only methods.

\paragraph{Hyperparameter Transferability.}
CRePE incurs an additional overhead of approximately 20 minutes for hyperparameter optimization via PHO. However, we find that this search need not be repeated for every target model. Table~\ref{tab:cross_sparsity} shows the results of directly applying hyperparameters optimized on LLaMA-2-7B---under unstructured, 2:4, and 4:8 sparsity settings---to LLaMA-2-13B. All three transferred configurations consistently outperform the RIA baseline across all sparsity patterns (unstructured: $5.83 \to 5.81$; 2:4: $8.41 \to 8.03$--$8.09$; 4:8: $6.74 \to 6.62$--$6.65$), demonstrating that PHO hyperparameters transfer well across model scales within the same family.

\begin{table}[hbt!]
\centering
\caption{Combination of CRePE with non-uniform sparsity allocation methods at 70\% unstructured.}
\label{tab:nonuniform}
\resizebox{\linewidth}{!}{%
\begin{tabular}{llccc}
\toprule
\multirow{2}{*}{\textbf{Method}} & \multirow{2}{*}{\textbf{Allocation}} & \multirow{2}{*}{\textbf{Distribution}} & \multicolumn{2}{c}{\textbf{LLaMA-1-7B}} \\
\cmidrule(lr){4-5}
 & & & \textbf{Sparsity} & \textbf{PPL} \\
\midrule
\multirow{4}{*}{Wanda}
 & -    & uniform     & 70\% & 251.19 \\
 & OWL   & non-uniform & 70\% & 24.29  \\
 & ALPHA & non-uniform & 70\% & 24.00  \\
 & DLP   & non-uniform & 70\% & 21.04  \\
\midrule
\multirow{4}{*}{RIA}
 & -    & uniform     & 70\% & 91.63  \\
 & OWL   & non-uniform & 70\% & 24.76  \\
 & ALPHA & non-uniform & 70\% & 24.07  \\
 & DLP   & non-uniform & 70\% & 20.99  \\
\midrule
\multirow{4}{*}{CRePE}
 & -    & uniform     & 70\% & 64.56  \\
 & OWL   & non-uniform & 70\% & 22.28  \\
 & ALPHA & non-uniform & 70\% & 23.07  \\
 & DLP   & non-uniform & 70\% & 19.23  \\
\bottomrule
\end{tabular}
}
\end{table}

\subsection{Combination with CRePE}

\subsubsection{Combination with Channel Permutation}


Table~\ref{tab:cp} presents the results of combining CRePE with Channel Permutation (CP) and LSA CP~\cite{zhang2024plug} under 2:4 and 4:8 sparsity settings.
Under the 2:4 setting, perplexity improves progressively from CRePE alone (10.74) to CRePE + CP (10.20) and CRePE + LSA CP (9.80).
A consistent trend is observed under the 4:8 setting: $8.16 \to 8.00 \to 7.85$.
Notably, CRePE combined with LSA CP achieves lower perplexity than RIA under the same configurations (2:4 + LSA CP: 10.03; 4:8 + LSA CP: 8.01), demonstrating that the importance scoring improvement of CRePE combines orthogonally with channel permutation techniques.

\subsubsection{Combination with Non-uniform Sparsity Allocation}

Table~\ref{tab:nonuniform} presents the results of combining CRePE with non-uniform layerwise sparsity allocation methods---OWL~\cite{yin2024owl}, AlphaPruning~\cite{lu2024alphapruning}, and DLP~\cite{chen2025dlp}---on LLaMA-1-7B at 70\% sparsity.
CRePE consistently outperforms RIA when combined with all three allocation strategies: OWL (22.28), AlphaPruning (23.07), and DLP (19.23). In particular, the CRePE + DLP combination achieves the lowest perplexity across all configurations.

\begin{table}[hbt!]
\centering
\small
\caption{Combination of CRePE with Re-pruning method. PPL on LLaMA-1-7B with unstructured.}
\label{tab:reprune_sparsity}
\begin{tabular}{lcc}
\toprule
\multirow{2}{*}{\textbf{Method}} & \multicolumn{2}{c}{\textbf{LLaMA-1-7B (unstructured)}} \\
\cmidrule(lr){2-3}
 & \textbf{60\%} & \textbf{70\%} \\
\midrule
Wanda              & 10.71 & 86.09 \\
Wanda + DSnoT      & 10.44 & 70.44 \\
RIA                & 10.41 & 86.07 \\
RIA + DSnoT        & 10.19 & 71.39 \\
CRePE             & 10.12 & 74.29 \\
CRePE + DSnoT     & 9.98  & 67.82 \\
\bottomrule
\end{tabular}
\end{table}

\subsubsection{Combination with Re-pruning}


Table~\ref{tab:reprune_sparsity} presents the results of combining CRePE with the DSnoT~\cite{zhang2023dsnot} re-pruning method across varying sparsity ratios on LLaMA-1-7B. At 60\% sparsity, CRePE + DSnoT achieves a perplexity of 9.98, the lowest among all combined methods. At 70\% sparsity, CRePE + DSnoT (67.82) likewise outperforms RIA + DSnoT (71.39), demonstrating that CRePE and DSnoT combine orthogonally and yield consistent gains across sparsity levels.
\section{Conclusion}
We presented CRePE, which incorporates 2D neighborhood context and adaptive coefficients into Relative Importance scoring, and PHO, a proxy-based framework that leverages Gini\textsubscript{pruned}--PPL correlation ($\rho = 0.953$) to reduce hyperparameter search from 11 hours to 20 minutes. Across LLaMA-1/2/3, Qwen, Phi-4, and DeepSeek under unstructured, 2:4, and 4:8 sparsity, CRePE consistently outperforms scoring-only baselines and combines orthogonally with Channel Permutation, non-uniform sparsity allocation, and re-pruning.

\section*{Limitations}
This work has several limitations. 
First, at extreme sparsity levels above 70\%, the performance gap between CRePE and RIA narrows compared to lower sparsity settings, suggesting that importance scoring alone becomes less effective at very high sparsity and complementary techniques such as weight  reconstruction may be necessary.
Second, while Gini\textsubscript{pruned} demonstrates strong empirical correlation with PPL ($\rho = 0.953$), a rigorous theoretical analysis of this relationship remains an open question for future work. 
Third, our method is limited to dense linear layers in LLMs; extending CRePE to structured components such as Mixture-of-Experts (MoE) routing mechanisms is left for future exploration. 
Finally, while CRePE inherits the efficiency of scoring-only methods, the impact of the additional convolution operation across diverse hardware environments has not been systematically evaluated.

\bibliography{custom}
\end{document}